\documentclass[10pt,journal,compsoc]{IEEEtran}
\usepackage{amsmath}
\usepackage{amssymb}
\usepackage{mathdots}
\usepackage{graphicx}
\usepackage{booktabs}
\usepackage{multirow}
\usepackage{tabularx}
\usepackage{titlecaps}
\ifCLASSOPTIONcompsoc
  \usepackage[nocompress]{cite}
\else
  \usepackage{cite}
\fi
\ifCLASSINFOpdf
\else
\fi
\begin{document}
\Resetlcwords
\Addlcwords{for a is but and with of in as the etc on to if by}
%
\title{Unitary Multi-Margin BERT \\for Robust Natural Language Processing}
%
%
%
%

\author{Hao-Yuan Chang
        and~Kang~L.~Wang
\IEEEcompsocitemizethanks{\IEEEcompsocthanksitem The authors are with the Department of Electrical and Computer Engineering, University of California, Los Angeles, 420 Westwood Blvd, Los Angeles, CA 90095, USA. E-mail: $\{$h.chang, klwang$\}$@ucla.edu\protect\\

}
\thanks{}}

%
%

\markboth{}%
{Chang \MakeLowercase{\textit{et al.}}: Unitary Multi-Margin BERT for Robust Natural Language Processing}
%



\IEEEtitleabstractindextext{%
\begin{abstract}
Recent developments in adversarial attacks on deep learning leave many mission-critical natural language processing (NLP) systems at risk of exploitation. To address the lack of computationally efficient adversarial defense methods, this paper reports a novel, universal technique that drastically improves the robustness of Bidirectional Encoder Representations from Transformers (BERT) by combining the unitary weights with the multi-margin loss. We discover that the marriage of these two simple ideas amplifies the protection against malicious interference. Our model, the unitary multi-margin BERT (UniBERT), boosts post-attack classification accuracies significantly by 5.3\% to 73.8\% while maintaining competitive pre-attack accuracies. Furthermore, the pre-attack and post-attack accuracy tradeoff can be adjusted via a single scalar parameter to best fit the design requirements for the target applications.
\end{abstract}

}

\maketitle

\IEEEdisplaynontitleabstractindextext

%
\IEEEpeerreviewmaketitle

\IEEEraisesectionheading{\section{Introduction}\label{sec:introduction}}

%
%
%
%
\IEEEPARstart{T}{he} objective of natural language processing (NLP) is to analyze human-generated texts with machines (e.g., automatic text classification). However, deep neural networks are prone to adversarial attacks. Without proper defense mechanisms in place, hackers can easily sabotage the prediction output of neural nets by maliciously perturbing the textual input, controlling the prediction in some cases. Such vulnerability prevents deep neural nets to be fully trusted in mission-critical applications such as transportation, national security, and defense. In this paper, we aim to design a robust machine learning model that can categorize the topic of a paragraph, understand the causal relationship between sentences, and infer the writer's sentiment while under adversarial attacks. To achieve this aim, we start by improving the state-of-the-art neural network in NLP: the \textbf{B}idirectional \textbf{E}ncoder \textbf{R}epresentations from \textbf{T}ransformers (\textbf{BERT})---a network structure designed specifically for understanding languages [1].

The BERT encodes words with real vectors, which are combined to embed the meaning of a sentence. To clarify our terminology used in this paper, the activation value of a neuron is defined as the sum of the products between the inputs and the synaptic weights, and a neural representation refers to the distribution of activations for a particular class of samples in a text classification task. In BERT, the subsequent attention layers transform the sentence embeddings into a succinct neural representation. For text classification particularly, the last two fully-connected layers serve as a classifier that categorizes the neural representations, and we call the activations of the last neural layer ``logits.'' During training, logits are normalized with the softmax function before passing to the cross-entropy loss in standard BERT models. During inference, BERTs use the argmax function to identify the corresponding class with the largest logit, resulting in an integer class label for the prediction. Although BERTs deliver excellent prediction accuracies, they can be easily attacked to produce incorrect outputs.

We define adversarial attacks as follows: an algorithm maliciously injects small perturbations into a neural network to alter its output prediction. The attackers aim to create just enough perturbation to shift the neural response across the decision boundary without appearing suspicious to humans [2]. Decision boundaries are the demarcation of data distributions for different classes in the activation space. The are many ways to devise small perturbations. Introducing typos or swapping synonyms are two popular methods, which will explain the detailed procedures in Sect. 5.2 later. 

Our innovations can be summarized into two integral parts as follows. Firstly, we discover that by switching the cross-entropy loss with the multi-margin loss as defined in Sect. 3.1 during the finetuning portion of the neural network training, our version of BERT forces the neural representation of different classes to be more distinct as we will show in Sect. 6.4. Better separation of the neural representations improves adversarial robustness, which is measured by the prediction accuracy under adversarial attacks (i.e., the post-attack accuracy) [3]. 

Secondly, deep neural nets have many sequential layers, and each layer has its own set of weight matrices. Normally, these matrices are unconstrained and free to take on any values, so we call them ``non-unitary weights.'' The problem with non-unitary weights is that they sometimes amplify the injected perturbation by accident, a vulnerability that the attackers can leverage to sabotage the prediction outputs. The current state-of-the-art BERT models have 12 to 24 attention layers with non-unitary weights, allowing small noises to be repeatedly amplified into large deviations. In addition to the aforementioned multi-margin loss, we propose another novel technique that uses unitary matrices to further boost the adversarial robustness of deep neural nets. Our proposed technique constrains the weight matrices to be unitary as we will explain in Sect. 3.2. Unitary weights maintain the cosine distance between the original and the perturbed sentence embedding vectors, hence, constraining the injected perturbations (proven later in Theorem 1). Unitarity counteracts the injected perturbations and stabilizes the final prediction results.

As a brief overview for this paper with detailed explanations to be followed, our scientific discoveries and contributions to the machine learning literature include:

\begin{enumerate}
\item  The use of the multi-margin loss to improve  BERT's adversarial robustness. The multi-margin loss encourages BERT to readjust its weights in the prior layers to better separate the neural representations of different classes; as a result, it provides a cushion to absorb the small input perturbations injected by the adversaries. 

\item  When used together with the multi-margin loss, the unitary weights further boost the adversarial robustness by constraining the injected perturbations across the network. 

\item  Our proposed neural network is attack-agnostic, simple, and unique. Unlike adversarial training, our model does not depend on the type of attacks used by the adversaries. Moreover, our model is much more straightforward to implement compared to the regularization methods.
\end{enumerate}
\noindent We name our innovation the \textbf{Uni}tary multi-margin \textbf{B}idirectional \textbf{E}ncoder \textbf{R}epresentations from \textbf{T}ransformers (\textbf{UniBERT}), which incorporates both the multi-margin loss and the unitary weights. UniBERT delivers significant improvements in post-attack accuracies compared to the state-of-the-art defense methods across multiple tasks as we will demonstrate in the following sections.

\section{Related Works}
To date, the techniques for defending against adversarial attacks on BERT can be grouped into two categories: 

\begin{enumerate}
\item  \textbf{Adversarial Training with Data Augmentation.} The first technique is to train the model with additional adversarial and augmented data. The designer of the neural network needs to anticipate the attack recipes that the hacker will use, generate adversarial samples accordingly, and train the network with the generated samples to prevent a future attack. The drawback of this technique is that it requires the designers to predict the attack methods used by the adversaries to create appropriate coverage in the sample space. Furthermore, training will take much longer if we desire full coverage for all types of adversarial attacks. Data augmentation is commonly used together with adversarial training, which generates additional training samples by swapping words with their synonyms or by interpolating between the existing sentence embedding vectors. One of the best is to combine adversarial training and data augmentation as seen in defense models such as AMDA [4] and MRAT [5] (see Sect. 6.2 for details). For brevity, we will use the term ``adversarial training'' to refer to defense models that use adversarial training and data augmentation together to achieve state-of-the-art robustness.

\item  \textbf{Regularization.} The second method of defense against adversarial attacks is by adding a regularization term in the loss function to reduce model complexity in particular ways. Normally, overfitting refers to the phenomenon in which the deep neural network learns all the artifacts in the training dataset, resulting in poor generalization accuracy when the model encounters new data points. We can view adversarial attacks as the generation of new, unseen data points outside of the training and the test datasets. In this sense, the original BERT overfits the training and test datasets, leading to poor accuracy on the adversarial samples. A way to alleviate this special type of overfitting is by regularizing the network to reduce the number of decision regions that contain the training data [6]. As another example, InfoBERT is a model that uses an information bottleneck as a regularizer to limit model complexity for improving adversarial robustness [7]. Disadvantages of regularization-based defense include high computational costs and added architectural complications. 
\end{enumerate}

\noindent To our knowledge, the multi-margin loss has not been studied for improving the robustness of NLP applications. Our study is the first time when the multi-margin loss and unitarity are used together to improve the adversarial robustness in NLP applications. The improvement is quite dramatic as shown in the following sections.

\section{Theory}
\subsection{Multi-margin Loss Increases Robustness}

\noindent Unlike most loss functions that only quantify the distance between the desired and the current output, the multi-margin loss provides an additional margin of safety between the logits and the decision boundary. The multi-margin loss is defined as: 
\begin{equation} \label{GrindEQ__1_} 
L\ \mathrm{\triangleq }\ \frac{\mathrm{1}}{n}\sum^{n\mathrm{-1}}_{i\mathrm{=0}}{\sum^{n_c\mathrm{-1}}_{j\mathrm{=0}}{{\mathrm{max} \left(y_{i,j}\ \mathrm{+}\ \varepsilon\  \mathrm{-}\ y_{i,correct},0\right)\ }}},  
\end{equation} 
where $n$ is the batch size, ${n}_{c}$ is the number of classes, ${y}_{i,correct}$ is the logit of the neuron corresponding to the correct answer, ${y}_{i,j}$ is the logit for other neurons; \textit{$\varepsilon$} is the margin parameter, setting a gap between the logits for different classes as discussed below. All variables are scalars. 

To provide some intuition for the multi-margin loss, we illustrate how it works in UniBERT with a binary sentiment analysis example in Fig. 1 below. The multi-margin loss is illustrated in the right panel of Fig. 1. For each data sample (indexed by \textit{i}) in \eqref{GrindEQ__1_}, the loss is the larger of the two terms between ${y}_{i,j}+\varepsilon-{y}_{i,correct}$ and zero, which correspond to the two linear segments of the graph in Fig. 1. The margin parameter (\textit{$\varepsilon$}) determines the intercept between the two segments. Our objective is to train the network such that the logit for the correct class (${y}_{i,correct}$) exceeds the logit of the incorrect class (${y}_{i,j}$) by the margin (\textit{$\varepsilon$}). In a support vector machine (SVM), the data points for the classifier are fixed; nevertheless, in UniBERT, the neural representation is modifiable during network training. A large \textit{$\varepsilon$} encourages our UniBERT to have a distinctive neural representation for each class. 

\begin{figure}[!t]
\centering
\includegraphics[width=3.5in]{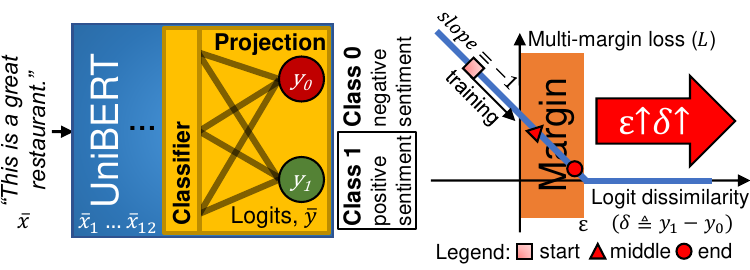}
\caption{Multi-margin loss for binary sentiment analysis with UniBERT. On the left, UniBERT receives an input sentence $\overline{x}$, transforming it with the embedding and the 12 attention layers (abbreviated as {\dots} in the figure) to latent neural representations, ${\overline{x}}_1\dots {\overline{x}}_{12}$. ${\overline{x}}_{12}$ passes through the classifier and the projection layers to become logits $\overline{y}$, consisting of $y_1$ and $y_0$ (scalars) in the binary classification example shown here. If $y_1>y_0$, the network will predict that the sentence has a positive sentiment (class 1); otherwise, it predicts a negative sentiment (class 0). The multi-margin loss compares $\overline{y}$ with the correct answer labeled by humans (denoted by the checkmark in the figure) and penalizes the network for any insufficient distinction between the two logits (i.e., $y_1-y_0$. We name this difference ``logit dissimilarity,'' \textit{$\delta$}. Additionally, we define a margin parameter \textit{$\varepsilon$} such that the multi-margin loss is proportional to the lack of the desired margin when \textit{$\delta$} $\mathrm{<}$ \textit{$\varepsilon$}  as shown in the figure by the line with a slope of -1. If the network has sufficient margin, the loss is zero (i.e., the flat segment on the right). During training, our UniBERT adjusts its weights from all the layers to minimize loss; training progression is depicted by the square, triangular, and circular markers on the graph. With a larger \textit{$\varepsilon$}, the multi-margin loss encourages our UniBERT to have highly distinctive logits, making it more difficult for the attackers to move them across the decision boundary with small perturbations and to sabotage the prediction results.}
\label{fig_1}
\end{figure}

Moreover, a more distinctive neural representation leads to higher adversarial robustness. Pang et al. [3] proved in theory that the amount of input perturbation required to cross the decision boundary monotonically increases with the Mahalanobis distance $d_M$ between the neural representations for the two classes. Assuming equal-variant Gaussian distributions, they compute the Mahalanobis distance as: 
\begin{equation} \label{GrindEQ__2_} 
d_M\ \mathrm{\triangleq }\ {\left({\overline{\mu }}_i\mathrm{-}{\overline{\mu }}_j\right)}^{\mathrm{T}}{\boldsymbol{S}}^{\mathrm{-1}}\left({\overline{\mu }}_i\mathrm{-}{\overline{\mu }}_j\right),   
\end{equation} 
where ${\overline{\mu }}_i$ and ${\overline{\mu }}_j$ and are the means of the neural representations for class \textit{i} and \textit{j}, respectively; $\boldsymbol{S}$ is the covariant matrix. In real-world applications, $\boldsymbol{S}$ can differ for the different classes, and the distribution of the neural representation may not be Gaussian. Despite this, we aim to confirm the following hypothesis experimentally: the multi-margin loss with a high margin parameter (\textit{$\varepsilon$}) forces UniBERT to separate its neural representations for different classes; thus, the distance between ${\overline{\mu }}_{\mathrm{i}}$ and ${\overline{\mu }}_{\mathrm{j}}$ increases. Given the same covariances, \textit{d${}_{M}$} in \eqref{GrindEQ__2_} should increase, requiring a larger input perturbation for the hackers to modify the prediction output. Consequently, the system is more resilient to adversarial attacks. We will show the effect of the multi-margin loss in Sect. 6.4 with experimental data.

\noindent 
\subsection{Unitarity Confines Perturbation}

\noindent In an adversarial attack, the input perturbation needs to be subtle to avoid detection. As mentioned previously, non-unitary weights can accidentally amplify the small perturbation, making the network vulnerable to adversarial attacks. Synaptic weights in a neural net are arranged in a matrix form. In the theorem below, we prove that a unitary weight matrix preserves the amount of the perturbation to the original sentence embedding, $\overline{x}$:

\newtheorem{theorem}{Theorem}
\begin{theorem} 
A unitary matrix ($\boldsymbol{U}$) maintains the Euclidean distance between the original ($\overline{x}$) and the perturbed vector ($\overline{x}^{'}$) after the linear transformation.
\end{theorem}
\begin{IEEEproof}
$
\|\boldsymbol{U}\overline{x}-\boldsymbol{U}{\overline{x}}^{'}\| 
= \sqrt{{{[\boldsymbol{U}(\overline{x}-\overline{x}^{'})]}^{T}\boldsymbol{U}(\overline{x}-\overline{x}^{'})}}
= \sqrt{{{(\overline{x}-\overline{x}^{'})}^{T}\boldsymbol{U}^{T}\boldsymbol{U}(\overline{x}-\overline{x}^{'})}}
= \|\overline{x}-{\overline{x}}^{'}\| 
$
\end{IEEEproof}

The right-hand side of the above equation is the L2 norm of the input perturbation, which will need to be small to be discreet. The left-hand side measures the amount of change in the output after the unitary transform. As shown in Theorem 1, unitarity guarantees that small perturbations remain small after the unitary transformation. In its non-unitary counterparts, the neural layers may accidentally amplify the perturbations, in which the logits can move across the decision boundary and result in a classification error. Non-unitary weights may suppress the perturbations as well; however, our goal is to eliminate any slight chance of perturbation amplification with unitary weights.

In our chosen implementation of the UniBERT, we impose unitary constraints on the selected layers to stabilize the injected perturbations across the neural network (see Sect. 4.2 for design details and discussion regarding non-linearity in the network). Sect. 6.5 confirms this stabilization effect of unitary weights by comparing the neural representations of the original and perturbed sentences across the network. Combining the multi-margin loss and the unitary weights, our UniBERT improves the state-of-the-art post-attack accuracies as we will show in Sect. 6.2.

\section{Model}
\subsection{Network Architecture \& Training}
\noindent Our UniBERT is an enhanced BERT that uses multi-margin loss and unitarity as described above to increase the neural network's prediction accuracy under adversarial attacks. This subsection explains the implementation details for our UniBERT neural architecture, which is illustrated in Fig. 2 below. Compared to the original BERT, our modifications are: 
\begin{enumerate}
\item During finetuning, we replace the softmax and cross-entropy loss with just the multi-margin loss on its own (see Fig. 2, bottom). We do not modify the softmax or the cross-entropy loss for pretraining. The reason for this is that the multi-margin loss works best for classification tasks; thus, it is applied during finetuning only and not during pretraining, which is a masked language modeling task. 

\item During both pretraining and finetuning, we force certain layers to have unitary weights, and the selected layers are circled with dash lines in Fig. 2, top-right. The selection of the unitary layers will be explained below in Sect. 4.2. The exact procedure to ensure their weight matrices are unitary is described in the next subsection. 
\end{enumerate}

\noindent Also, although our current implementation has the same number of parameters (110M) as the BERT base, researchers have shown that it is possible to further compress the number of trainable parameters by half [8]. Detailed training procedures are described in Appendix A for reproducibility. We set the new margin parameter ($\varepsilon$) in \eqref{GrindEQ__1_} to 100 for balanced pre-attack and post-attack accuracies. In general, a larger $\varepsilon$ will slightly lower the pre-attack accuracy while boosting the post-attack accuracy. See Appendix B for details on the selection of the margin parameter ($\varepsilon$) via hyperparameter tuning and the pre-attack vs. post-attack accuracies tradeoff. Our source code can be downloaded from https://github.com/h-chang/UniBERT.

\begin{figure}[!t]
\centering
\includegraphics[width=3.5in]{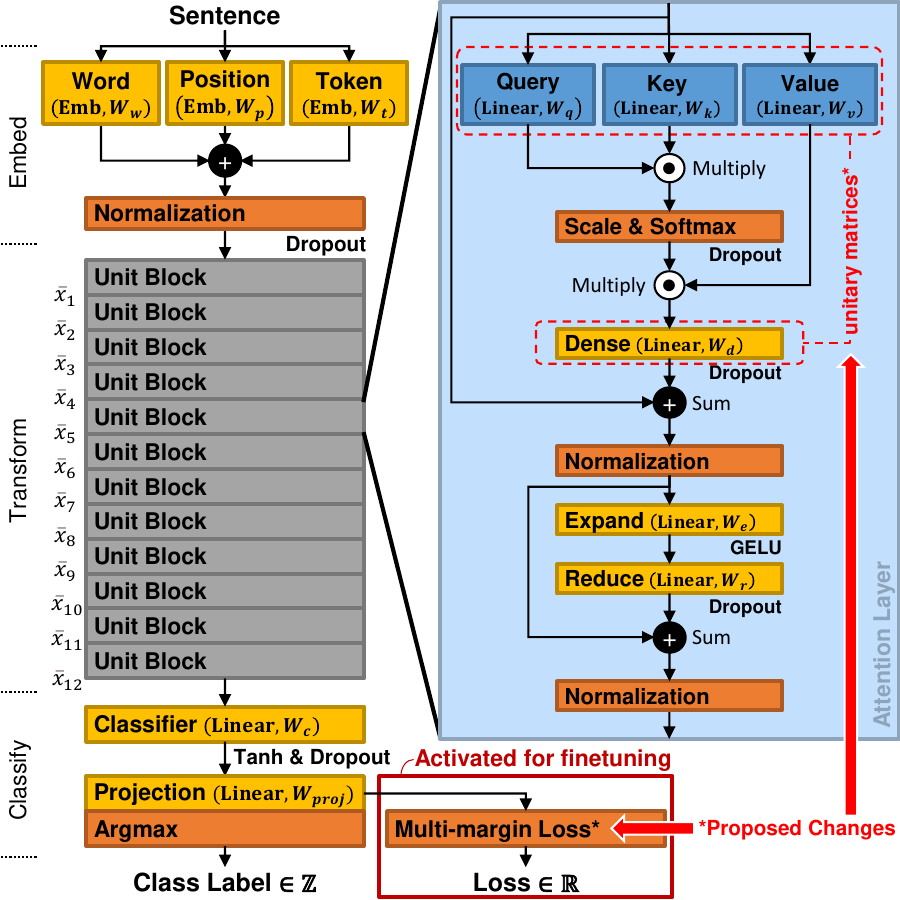}
\caption{Our unitary multi-margin BERT (UniBERT) architecture. Like BERT, UniBERT classifies a sentence by first converting words in a sentence vector using the Word, Position, and Token embeddings (top-left). It then transforms this sentence vector using 12 attention layers with details shown on the right of this figure. Our UniBERT is a variant of BERT with the following differences: First, we use the multi-margin loss (bottom) instead of the cross-entropy loss during the finetuning portion of the training process. Second, we enforce unitary constraints on the weights circled with dashed lines (top-right). These enhancements increase the resilience to input perturbations for stabilizing the classification outcomes under adversarial attacks. ${\overline{x}}_1\dots {\overline{x}}_{12}$ denotes the activations after each of the 12 unit blocks, respectively (left). Emb means an embedding layer while Linear means a fully-connected layer. Tanh is the hyperbolic tangent; GELU is the Gaussian Error Linear Unit. They are nonlinearities similar to the ReLU. }
\label{fig_2}
\end{figure}

\subsection{Selection of the Unitary Layers}

\noindent The unitary layers are selected as follows, and Table 1 below lists all synaptic weights in UniBERT along with their unitarity. Firstly, only square matrices can be unitary in terms of a strict mathematical definition; hence, not all layers can be converted to unitary. We decide to apply unitary constraints on all the square weights in UniBERT to maximize the effect of the total unitarity (with one exception to be explained next). Secondly, we keep \textbf{\textit{W${}_{c}$}} in the classifier layer non-unitary to increase the pre-attack accuracy (i.e., prediction accuracy without adversarial attacks), even though it is also square and thus available for the unitary conversion. \textbf{\textit{W${}_{c}$}} is shown near the bottom of Fig. 2 and listed in Table 1. In general, unitarity reduces model complexity in exchange for higher signal stability [8]. This is the reason why with unitary constraints, the pre-attack accuracies decrease (usually by a small amount), and the post-attack accuracies increase. As a consequence, we decide to leave the classifier layer non-unitary for a slightly higher pre-attack accuracy.

\begin{table}[!t]
\renewcommand{\arraystretch}{1.3}
\caption{\titlecap{The complete list of weights used in our UniBERT and their unitarity}}
\label{table_1}
\centering
\begin{tabular*}{\columnwidth}{lp{0.3in}lccc}
\hline 
\textbf{Name} & \textbf{Symbol} & \textbf{Module} & \textbf{Dimension} & \textbf{Unitary} & \textbf{Repeat} \\ 
\hline 
Word & \textbf{\textit{~W${}_{w}$}} & Embed & 30522 x 768 & No & 1 \\ 
Position & \textbf{\textit{~W${}_{p}$}} & Embed & 512 x 768 & No & 1 \\ 
Token & \textbf{\textit{~W${}_{t}$}} & Embed & 2 x 768 & No & 1 \\ 
\textit{Query} & \textbf{\textit{~W${}_{q}$}} & \textit{Linear} & \textit{768 x 768} & \textit{Yes} & \textit{12} \\ 
\textit{Key} & \textbf{\textit{~W${}_{k}$}} & \textit{Linear} & \textit{768 x 768} & \textit{Yes} & \textit{12} \\ 
\textit{Value} & \textbf{\textit{~W${}_{v}$}} & \textit{Linear} & \textit{768 x 768} & \textit{Yes} & \textit{12} \\ 
\textit{Dense} & \textbf{\textit{~W${}_{d}$}} & \textit{Linear} & \textit{768 x 768} & \textit{Yes} & \textit{12} \\ 
Expand & \textbf{\textit{~W${}_{e}$}} & Linear & 768 x 3072 & No & 12 \\
Reduce & \textbf{\textit{~W${}_{r}$}} & Linear & 3072 x 768 & No & 12 \\ 
Classifier & \textbf{\textit{~W${}_{c}$}} & Linear & 768 x 768 & No & 1 \\ 
Projection & \textbf{\textit{~W${}_{proj}$}} & Linear & 768 x \textit{n${}_{c}$} & No & 1 \\ 
\hline 
\end{tabular*}
\raggedright
\begin{footnotesize}
\newline
Each row in the table is a neural layer with a corresponding weight matrix. We force unitary constraints on all the square weights except the classifier layer (i.e., \textbf{\textit{W${}_{c}$}}), which we leave unconstrained to achieve higher pre-attack accuracy. Also, non-square weights are difficult to make unitary and therefore left as is. In the projection layer, ${n}_{c}$ is the number of classes in the classification task. Italic rows are the unitary weights used in UniBERT; they are non-unitary in the original BERT.
\end{footnotesize}
\end{table}

Our UniBERT is only partially unitary as shown in Table 1. It has a total of 48 unitary weights and 29 non-unitary weights (Table 1, last two columns). Our selection of unitary layers forces unitary constraints on 62.3\% of all weight matrices; i.e., 48 / (48 + 29). Likewise, there are nonlinear activation functions in the network necessary to achieve the required model complexity. The nonlinearities are vital to avoid condensing the transfer function to one linear transformation between the input and the output. While it is impossible to have a purely unitary neural net, our UniBERT's partial unitarity is sufficient to deter adversarial attacks, and surprisingly, it outperforms many state-of-the-art defense models in post-attack accuracies as we will demonstrate in Sect. 6.2.

\subsection{Unitary Constraints}

\noindent The way we convert the non-unitary weights to their closest unitary projections is by QR factorization, a method to decompose any matrix into unitary and non-unitary parts as explained as follows [9]:
\begin{equation} \label{GrindEQ__4_} 
\boldsymbol{W}\ \mathrm{=}\ \boldsymbol{QR},  
\end{equation} 
where $\boldsymbol{W}$ is a non-unitary square matrix, $\boldsymbol{Q}$ is a unitary matrix, and $\boldsymbol{R}$ is an upper triangular matrix. We extract the signs of the diagonal elements in $\boldsymbol{R}$ and construct $\boldsymbol{S}$: 
\begin{equation} \label{GrindEQ__5_} 
\boldsymbol{S}\ \mathrm{=}\ diag\mathrm{(}sign\left(\boldsymbol{R}\right)\mathrm{)} 
\end{equation} 
Therefore, $\boldsymbol{S}$ is a diagonal matrix with $\mathrm{\pm}$1 on its diagonal. Lastly, we obtain the unitary weights ($\boldsymbol{U}$) by: 
\begin{equation} \label{GrindEQ__6_} 
\boldsymbol{U}\ \mathrm{=}\ \boldsymbol{QS} 
\end{equation} 

$\boldsymbol{U}$ is proven to be unitary because:
\begin{equation} \label{GrindEQ__7_} 
{\boldsymbol{U}}^T\boldsymbol{U}\ \mathrm{=}\ {\boldsymbol{S}}^T{\boldsymbol{Q}}^T\boldsymbol{QS}\ \mathrm{=}\ {\boldsymbol{S}}^T\boldsymbol{S}\ \mathrm{=}\ \boldsymbol{I}
\end{equation} 

\noindent After each backpropagation step in neural network training, we use the procedure detailed in \eqref{GrindEQ__4_}-\eqref{GrindEQ__6_} to find the closest unitary projection for selected weights and overwrite the weights with $\boldsymbol{U}$.

\section{Experiments}
\subsection{Datasets}
\noindent Here we report the datasets used in pretraining and finetuning to benchmark the differences between BERT and UniBERT. First, we describe the dataset used for pretraining---the Book Corpus (\textbf{bookcorpus}), an unlabeled dataset containing 74 million sentences from eleven thousand books [10]. We separate this dataset into two subsets, one for training (95\%) and one for testing (5\%). Then for finetuning, we selected three different datasets for a comprehensive evaluation covering multilabel categorization, language inference, and sentiment analysis; respectively, they are listed as follows:

\begin{enumerate}
\item  AG's News (\textbf{ag\_news}) is a dataset for news classification, and the goal is to classify the articles into four categories including world news, business news, science \& technology, and sports [11]. 

\item  Stanford Natural Language Inference Corpus (\textbf{snli}) aims to train machine learning systems that can identify the relationship between a pair of sentences [12]. There are three possible classification outcomes: entailment, contradiction, or neutral. 

\item  Yelp Reviews Polarity (\textbf{yelp}) is a text sentiment analysis dataset constructed by collecting reviews from Yelp.com [11]. The label is either positive or negative. 
\end{enumerate}

\noindent We highlight the key features of the dataset statistics in Appendix C.

\noindent 
\subsection{Adversarial Attacks}

\noindent Attackers may swap the words in a sentence with their synonyms to cause a misclassification, and we refer to this type of attack as synonym-based attacks [13], [14]. Another type of adversarial attack is created by introducing typographical errors in the sentences [15], [16]. In this case, although the injected typos appear benevolent to the readers, they can manipulate BERT to produce the wrong results because non-unitary weights may amplify the small perturbations. Below lists the NLP attacks we evaluate in this study. We select three distinct types of attacks for a comprehensive adversarial robustness evaluation:

\begin{enumerate}
\item  \textbf{Textbugger} randomly introduces character insertion, deletion, swap, and substitution to modify BERT's prediction [16]. It is considered a typographic attack.

\item  \textbf{Textfooler} finds candidate adversarial samples by swapping important words with their synonyms; synonyms are found by searching for the closest words in the word embedding space [14], [17]. In particular, the counter-fitted word embeddings are used for the synonym search.

\item  \textbf{Probability Weighted Word Saliency} (or \textbf{PWWS}) [18] swaps words in a sentence with their synonyms as defined in the human-labeled WordNet [19] database. In contrast to Textfooler, which finds synonyms using a distance metric with the word embeddings learned automatically by a neural net, PWWS relies on a thesaurus constructed explicitly by human workers.
\end{enumerate}

Textbugger and Textfooler may not preserve a sentence's meaning; therefore, to create a valid replacement sentence to commence the attack, we need to check for the semantic similarity between the generated adversarial sample and the original sentence. We use the Universal Sentence Encoder [20] to measure the cosine distance between the original and the perturbed sentences and use the default similarity threshold in the TextAttack [21] framework${}^{\ }$to reject any adversarial example that changes the meaning of the sentence. In contrast, because PWWS uses the human-labeled WordNet synonym database to generate high-fidelity samples as described before; hence, we do not perform additional safeguarding on the generated samples. 

Furthermore, there are two ways to carry out adversarial attacks on a neural network: \textit{targeted attacks} vs. \textit{static attacks} [4]. Targeted attacks generate a new set of adversarial samples for each neural network while static attacks use the same set of adversarial samples to evaluate all neural network architectures (e.g., BERT, UniBERT, {\dots}etc.). We use the tougher targeted attacks to evaluate the model's robustness. In detail, we randomly select 1000 data samples from the test set and allow the attack algorithm to make an unlimited number of attempts to the model until it can no longer generate new permutations that meet the similarity criteria or have exhausted all synonyms. The number of attempts the attacker is allowed to make is called the \textit{query budget}. The \textit{post-attack accuracy} is calculated as the ratio of the samples (out of 1000) that survive the series of attacks with an unlimited query budget and still produce the correct classification results, measuring the classification accuracy of the neural network under adversarial attack. Our evaluation methodology is the toughest in the literature, reporting the lowest possible post-attack accuracies. The \textit{pre-attack accuracy} is measured by computing the ratio of the correctly classified sample out of the 1000 test samples without any attack.

\section{Result \& Discussion}
\subsection{Our UniBERT vs. Baseline Models}
\noindent So far, we have only introduced BERT and UniBERT, but there are many versions of BERT for various purposes. These neural network models are not specifically designed to defend against adversarial attacks; as a result, we call them the baseline models. The goal is to validate the need for a novel defense technique because there would be no reason to modify the existing architectures if the baseline models were already resilient to attacks. We compare our proposed UniBERT against four other versions of BERT, which are listed here:

\begin{enumerate}
\item  \textbf{B}idirectional \textbf{E}ncoder \textbf{R}epresentations from \textbf{T}ransformers (\textbf{BERT}) [1]---This is the original architecture that our work is based on. 

\item  A \textbf{R}obustly \textbf{O}ptimized \textbf{BERT} Pretraining \textbf{A}pproach (\textbf{RoBERTa}) [22]---This is a BERT with enhanced pretraining to improve its accuracy. We use this model to verify that a better pretraining procedure alone cannot deter adversarial attacks.

\item  \textbf{A} \textbf{L}ite \textbf{BERT }(\textbf{ALBERT})\textbf{ }[23]---This model reduces the number of parameters up to 18 times by using factorized word embedding and parameter sharing. We select this model to show that a simple model complexity reduction is not enough to prevent adversarial attacks.

\item  A \textbf{distil}led version of \textbf{BERT }(\textbf{DistilBERT})\textbf{ }[24]---This model reduces the number of parameters by 40\% by using knowledge distillation. Same as with ALBERT, we want to show that reducing the model complexity via transfer learning to a smaller model will not improve robustness.
\end{enumerate}

To conduct a fair comparison for the classification accuracy, we download the pretrained weights for the four baseline models above from the Hugging Face repository [25] and finetune them further using the procedure for our selected datasets. Namely, the classification datasets include news categorization (ag\_news), natural language inference (snli), and sentiment analysis (yelp), which we have documented in Sect. 5.1 above. After finetuning, we first evaluate their pre-attack classification accuracies and then measure their post-attack classification accuracies after performing adversarial attacks on the models. The three attack recipes are the PWWS, Textbugger, and Textfooler adversarial attacks on the textual input, which cover a wide range of typographical and synonym-based NLP attacks. Details of the attack recipes are discussed in Sect. 5.2 before.

The results are given in Table 2, which compares the pre-attack and post-attack accuracies of our proposed UniBERT with the baseline models (i.e., BERT, RoBERTa, ALBERT, and DistilBERT). Our observations regarding the four baseline models are as follows: RoBERTa generally delivers slightly better pre- and post-attack accuracies compared to BERT because RoBERTa's extensive pretraining creates a better language model. On the other hand, both parameter reduction models (ALBERT and DistilBERT) have the worse robustness against adversarial attacks. We conjecture that the lack of model complexity prohibits them to generalize to out-of-distribution samples due to their inferior language models. 

UniBERT delivers double-digit improvements in post-attack accuracies across all combinations of NLP tasks and attacks over the baseline models. The amounts of improvements in post-attack accuracies vary between 13.7\% (UniBERT vs. RoBERTa for snli under Textbugger) and 74.5\% (UniBERT vs. ALBERT for yelp under Textfooler). These data confirm our hypothesis that the combination of the multi-margin loss and unitarity improves robustness. Nevertheless, as a side effect, the unitary constraint also reduces model complexity. As a review of Sect. 4.2 earlier, the number of parameters is about half of a fully connected layer when we count a real, adjustable scalar number as one parameter. Specifically, the number of free parameters is \textit{n} (\textit{n }-- 1) / 2 for the orthogonal group of degree \textit{n}, matching the number of Lie algebras that the group has [8]. Consequently, we only need half as many real scalars to specify a neural layer under the unitary constraint. Because of the reduction in model complexity, UniBERT's pre-attack accuracy is closer to the reduced-size models (e.g., ALBERT and DistilBERT) as supposed to the full-size models (e.g., BERT). Despite a small degradation in the pre-attack accuracies, the UniBERT's strong suit is in post-attack accuracies, at which it surpasses all models with large margins. 

Neither the RoBERTa's enhanced pretraining nor ALBERT's or DistilBERT's model reduction results in significant improvements in the post-attack accuracy. With the data listed in Table 2, we confirm that adversarial attacks cannot be prevented with the baseline models alone. Further interventions are needed to counter the adversarial attacks. As an example of such intervention, in UniBERT, we combine the multi-margin loss with unitary weights to deliver superior adversarial robustness, and we will compare our model with other state-of-the-art defense methods next.

\begin{table}[!t]
\renewcommand{\arraystretch}{1.3}
\caption{\titlecap{Classification accuracies of our UniBERT vs. other baseline BERT models under attacks}}
\label{table_2}
\centering
\begin{tabularx}{\columnwidth}{p{0.1in}r>{\centering}p{0.7in}p{0.3in}p{0.45in}p{0.3in}} \hline 
\textbf{Task} & \textbf{Model} & \multirow{2}{*}{\shortstack{\textbf{Pre-attack}\\\textbf{Accuracy (\%)}}} & \multicolumn{3}{c}{\textbf{Post-attack Accuracy (\%)}} \\
 & & & PWWS & Textbugger & Textfooler \\ \hline 
\multirow{5}{*}{\rotatebox[origin=c]{90}{ag\_news}} & BERT & 94.6 & 32.4 & 35.8 & 13.4 \\ 
 & RoBERTa & \textbf{95.1} & 40.8 & 48.3 & 16.7 \\ 
 & ALBERT & 92.5 & 24.1 & 17.3 & 8.7 \\ 
 & DistilBERT & 93.6 & 27.9 & 32.6 & 10.6 \\ 
 & \textbf{UniBERT} & 92.3 & \textbf{85.4} & \textbf{83.1} & \textbf{82.3} \\ \hline
\multirow{5}{*}{\rotatebox[origin=c]{90}{snli}} & BERT & 90.1 & 1.5 & 4.0 & 3.8 \\ 
 & RoBERTa & \textbf{91.2} & 1.3 & 5.0 & 4.0 \\ 
 & ALBERT & 87.6 & 0.2 & 1.5 & 1.8 \\ 
 & DistilBERT & 88.2 & 0.8 & 2.5 & 2.9 \\ 
 & \textbf{UniBERT} & 86.6 & \textbf{21.5} & \textbf{18.7} & \textbf{17.7} \\ \hline 
\multirow{5}{*}{\rotatebox[origin=c]{90}{yelp}} & BERT & 95.4 & 3.9 & 15.9 & 2.9 \\ 
 & RoBERTa & \textbf{96.6} & 8.4 & 24.6 & 8.4 \\ 
 & ALBERT & 94.7 & 1.3 & 5.9 & 1.4 \\ 
 & DistilBERT & 95.9 & 4.9 & 17.4 & 2.6 \\  
 & \textbf{UniBERT} & 93.3 & \textbf{75.3} & \textbf{79.6} & \textbf{75.9} \\ \hline 
\end{tabularx}
\raggedright
\begin{footnotesize}
\newline
We study the typographical (PWWS), typographical \& synonym-based (Textbugger), and synonym-based (Textfooler) attacks on news categorization (ag\_news), natural language inference (snli), and sentiment analysis (yelp) for a comprehensive evaluation on text classification robustness. In the table, we compare the post-attack accuracies of our UniBERT against four other BERT variants (namely, BERT, RoBERTa, ALBERT, and DistilBERT), showing that our UniBERT outperforms the baseline models by a large margin. Accuracy is reported as a percentage (out of 100) to measure the ratio of correct prediction. The best accuracies are bolded.
\end{footnotesize}
\end{table}

\subsection{Our UniBERT vs. Defense Models}

\noindent We compare UniBERT's performance with the state-of-the-art defense techniques (e.g., adversarial training and regularization as outlined previously in Sect. 2), which we refer to as the ``defense models.'' We select the following representative defense models for comparison:

\begin{enumerate}
\item  \textbf{A}dversarial and \textbf{M}ixup \textbf{D}ata \textbf{A}ugmentation\textbf{ }(\textbf{AMDA-Tmix} \& \textbf{AMDA-Smix}) [4]---In this technique, adversarial examples are generated by assuming a specific attack recipe, and extra training data are created by linearly interpolating between the neural representations for different classes. They designate their model with the ``Tmix'' and ``Smix'' postfixes to denote the location where the neural representations are taken from in BERT. 

\item  \textbf{M}ixup \textbf{R}egularized \textbf{A}dversarial \textbf{T}raining (\textbf{MRAT }\& \textbf{MRAT+}) [5]---This technique uses all the steps detailed in AMDA above. In addition, it adds a regularization term in the loss function (i.e., a penalty for any data point that is too different from the original), ensuring that the augmented data will follow the original data distribution. They claim that this term will prevent the augmented dataset from degrading the pre-attack accuracy. Like AMDA, the key ingredient for improving robustness still comes from adversarial training with data augmentation. The MRAT+ variant adds data augmentation to the original data. For example, it may swap some words in the sentence with their synonyms to generate new samples.

\item  \textbf{Info}rmation Bottleneck on \textbf{BERT} (\textbf{InfoBERT}) [7]---This technique reduces model complexity by using an information bottleneck. The bottleneck is implemented as two regularization terms in the loss function: one is to maximize the prediction accuracy while minimizing the mutual information between the input and the internal representation. Another is to identify word embeddings that are less affected by the input perturbation and force the neural net to utilize these words more in its decision process. InfoBERT is computationally expensive due to the need to calculate mutual information. It also requires more hyperparameters, which makes model optimization cumbersome. 
\end{enumerate}

As explained in Sect. 5.2 previously, there are two ways to measure the post-attack classification accuracy: with targeted attacks or with static attacks. Thus, we need to make sure that we use the same evaluation method to compare the adversarial robustness of different models. The authors of AMDA and MRAT have published the performance of their models using targeted attacks, and their data are reproduced in the table below for comparison. The creator of InfoBERT uses static attacks in their paper; thus, we rerun their simulations with the targeted attacks for a fair comparison. The adversarial performance of our UniBERT is always evaluated with the targeted attacks. We first study the performance of the state-of-the-art defense models by comparing the data between Table 2 and Table 3. Defense models utilizing adversarial training (i.e., AMDA-Tmix/Smix and MRAT/+) deliver superior adversarial performance over BERT and RoBERTa. On the other hand, regularization-based models such as InfoBERT only deliver marginal improvements over the classic BERT model under the targeted attacks. Moreover, InfoBERT's robustness is slightly higher than RoBERTa's in natural language inference (snli), but it's worse in news categorization (ag\_news) and sentiment analysis (yelp). 

\begin{table}[!t]
\renewcommand{\arraystretch}{1.3}
\caption{\titlecap{Classification accuracies of our UniBERT vs. state-of-the-art defense models}}
\label{table_3}
\centering
\begin{tabularx}{\columnwidth}{p{0.1in}r>{\centering}p{0.57in}p{0.3in}p{0.45in}p{0.3in}} \hline 
\textbf{Task} & \textbf{Model} & \multirow{2}{*}{\shortstack{\textbf{Pre-attack}\\\textbf{Accuracy (\%)}}} & \multicolumn{3}{c}{\textbf{Post-attack Accuracy (\%)}} \\
 &  &  & PWWS & Textbugger & Textfooler \\ \hline 
\multirow{4}{*}{\rotatebox[origin=c]{90}{ag\_news}}  & AMDA-Tmix & \textbf{94.5} & 69.7 & N/R & 56.3 \\ 
 & AMDA-Smix & 94.3 & 70.0 & N/R & 51.3 \\ 
 & InfoBERT & 94.4 & 35.5 & 46.0 & 12.9 \\ 
 & \textbf{UniBERT} & 92.3 & \textbf{85.4} & \textbf{83.1} & \textbf{82.3} \\ \hline 
\multirow{4}{*}{\rotatebox[origin=c]{90}{snli}} & MRAT & 89.5 & N/R & 9.9 & 10.5 \\ 
 & MRAT+ & 88.7 & N/R & 12.2 & 12.4 \\ 
 & InfoBERT & \textbf{90.9} & 2.8 & 5.5 & 5.4 \\ 
 & \textbf{UniBERT} & 86.6 & \textbf{21.5} & \textbf{18.7} & \textbf{17.7} \\ \hline 
\multirow{2}{*}{\rotatebox[origin=c]{90}{yelp}} & InfoBERT & \textbf{97.4} & 3.9 & 21.0 & 2.1 \\ 
 & \textbf{UniBERT} & 93.3 & \textbf{75.3} & \textbf{79.6} & \textbf{75.9} \\ \hline 
\end{tabularx}
\raggedright
\begin{footnotesize}
\newline
AMDA-Tmix, AMDA-Smix [4], MRAT, and MRAT+ [5] use adversarial training with mix-up data augmentation to deliver the previous state-of-the-art performance, and their data are reproduced here. InfoBERT uses information bottleneck as a regularizer to slightly improve the post-attack accuracies. Out of all the defense techniques, UniBERT gives much higher post-attack accuracies over all attack recipes (PWWS, Textbugger, Textfooler). Accuracy is reported as a percentage (out of 100) to measure the ratio of correct prediction, and N/R indicates not reported in the original publication. The best accuracies are bolded.
\end{footnotesize}
\end{table}

Our UniBERT outperforms other state-of-the-art defense models in post-attack accuracies across all tasks and attack recipes. The enhancements in post-attack accuracies range from 5.3\% (UniBERT vs. MRAT+ for snli under Textfooler) to 73.8\% (UniBERT vs. InfoBERT for yelp under Textfooler). More specifically, in terms of post-attack accuracies, our UniBERT is 15.4\% to 31.0\% better in ag\_news and 5.3\% to 8.8\% better in snli compared to adversarial training. We explain the improvements in adversarial robustness between UniBERT and adversarial training as follows: Under a targeted attack with an unlimited query budget, there is a wide variety of perturbations an attack recipe can produce for a given sentence. Although adversarial training captures a decent portion of these variations with augmented data, it is impossible to be comprehensive and cover the entire search space. Instead of training adversarial samples by brute force, our UniBERT mitigates the vulnerability fundamentally by increasing the neural representations' distances to the decision boundaries (as explained later in Sect. 6.4) while confining the injected perturbations (Sect. 6.5). 

In UniBERT, the combination of these two innovations (i.e., multi-margin loss and unitarity) not only improves the robustness beyond the state-of-the-art but also provides a model that is attack-agnostic. The robustness enhancement is consistent across various attack recipes (i.e., PWWS, Textbugger, and Textfooler) as shown in the post-attack accuracy columns in Table 3. For UniBERT, the largest variation in post-attack accuracies is between 3.1\% and 4.3\% across different attack recipes. Other models are less consistent when compared across attacks. As an example, InfoBERT has a 33.1\% variation in post-attack accuracies across different attacks for ag\_news and AMDA-Smix has 18.7\%, but for snli, their variations are much smaller. Such inconsistency will cause the designers more time in selecting the best defense model. In summary, the key weakness of adversarial training is that it requires the designers to anticipate the types of attacks and create appropriate coverage, which is not always possible. And the advantage of UniBERT is that it does not assume which attack the attackers will use. 

Regularization methods such as InfoBERT perform poorly under targeted attacks with unlimited budgets, which is extremely aggressive in finding adversarial examples. Although InfoBERT is effective in preventing static attacks, we observe that under our aggressive, targeted attacks, InfoBERT is insufficient for improving adversarial robustness, evident by its poor post-attack accuracies in Table 3. However, InfoBERT delivers superior pre-attack accuracies that often exceed that of BERT in our observation. In terms of the pre-attack accuracies, our UniBERT has a slight degradation for reasons explained in the previous subsection but is still on par with adversarial training (namely, AMDA-Smix and MRAT+). 

\noindent 
\subsection{Ablation Study}

\noindent To separate the contribution of the multi-margin loss and the unitary weights, we perform an ablation study for UniBERT on the yelp task. Table 4 reports the performance of the following four model trims: 

\begin{enumerate}
\item  \textbf{BERT} is the baseline with no modification.

\item  \textbf{BERT\_unitarity} is BERT with the unitary constraints alone.

\item  \textbf{BERT\_multi\_margin} is BERT with the multi-margin loss instead of the cross-entropy loss without any unitary constraint.

\item  \textbf{UniBERT }has both the unitary weights and the multi-margin loss replacement.
\end{enumerate}

BERT\_unitarity's post-attack accuracies are not significantly different from BERT (Table 4, first two rows). The reason is the following: A query budget is defined as the number of attempts the attacker is allowed to make before giving up. In the case of adversarial attacks with infinite query budgets, the attackers are free to try as many permutations as possible until they exhausted all combinations. Because the post-attack accuracies reported here are measured with attacks with infinite query budgets, the attacker has a large search space to find one adversarial example that changes the prediction outcome, and it only needs one example for the attack to be considered as successful. If the original sentence's neural representation is close to the decision boundary in any direction in the high-dimensional embedding space, our aggressive attack procedure will find an adversarial sample in its proximity with high probability. Consequently, we do not see improvement with unitarity alone.

 BERT\_multi\_margin delivers a significant improvement in the post-attack accuracy but also reduces the pre-attack accuracy. As we discussed earlier, the multi-margin loss increases the distance to the decision boundaries, making it harder for the injected perturbation to cause a misclassification. Thus, we observe a significant increase (47.3\%-59.1\%) in post-attack accuracies. Even with increased distance to the decision boundaries, the non-unitary weights used in BERT\_multi\_margin can still amplify injected perturbations. By adding the unitary constraints on top of BERT\_multi\_margin, UniBERT raises both the pre-attack and post-attack accuracies considerably. Comparing to BERT\_multi\_margin, UniBERT boost the post-attack accuracies by an additional 12.3\% to 21.1\% via adding the unitary constraints on the synaptic weights. With this ablation study, we conclude that it is insufficient to have unitarity or multi-margin loss alone; the multi-margin loss and unitarity need to be used together to achieve optimal results. 

\begin{table}[!t]
\renewcommand{\arraystretch}{1.3}
\caption{\titlecap{Ablation study to understand the contribution of each component}}
\label{table_4}
\centering
\begin{tabularx}{\columnwidth}{p{0.9in}>{\centering}p{0.57in}p{0.3in}p{0.45in}p{0.3in}} \hline 
 \textbf{Model} & \multirow{2}{*}{\shortstack{\textbf{Pre-attack}\\\textbf{Accuracy (\%)}}} & \multicolumn{3}{c}{\textbf{Post-attack Accuracy (\%)}} \\
 & & PWWS & Textbugger & Textfooler \\ \hline 
BERT & \textbf{95.4} & 3.9 & 15.9 & 2.9 \\ \hline 
BERT\_unitarity  & 95.9 & 3.1 & 14.1 & 1.8 \\ \hline 
BERT\_multi\_margin & 82.9 & 63.0 & 63.2 & 54.8 \\ \hline 
UniBERT & 93.3 & \textbf{75.3} & \textbf{79.6} & \textbf{75.9} \\ \hline 
\end{tabularx}
\raggedright
\begin{footnotesize}
\newline
BERT is the baseline model. BERT\_unitarity adds unitary constraints to BERT without the multi-margin loss. BERT\_multi\_margin replaces the loss with multi-margin without placing any unitary constraints. This data shows that we need to use the multi-margin loss together with unitarity to achieve the best post-attack accuracies in UniBERT. The multi-margin loss is used for encouraging distinct neural representations while the unitarity is for stabilizing perturbations. Accuracy is reported as a percentage (out of 100) to measure the ratio of correct prediction. The best accuracies are bolded. We use the yelp dataset for the ablation study.
\end{footnotesize}
\end{table}

\noindent 
\subsection{Effect of the Multi-margin Loss}

\noindent Previously in Sect 3.1, we hypothesize that the multi-margin loss makes the neural representations more distinct. To confirm this, we record the logits (i.e., the neural representations at the last layer) in our UniBERT for 950 correctly labeled data samples with the yelp binary classification task, compared with the ones in BERT. With the yelp dataset, our model is asked to categorize the input text into a positive or a negative review (i.e., two classes). Samples with the wrong prediction outcome are ignored because the attack algorithms skip any misclassified sentences, so they will never be attacked and will not affect the robustness evaluation. For each data sample, the shortest distance to the decision boundary (\textit{d}) is calculated. Then, we define a new quantity called normalized distance, \textit{d${}_{s}$}, to measure the size of the buffer for absorbing the adversarial perturbation to prevent misclassification: 
\begin{equation} \label{GrindEQ__8_} 
d_s\ \mathrm{\triangleq }\ \mathrm{Mean(}d)\ /\ \mathrm{Var}(d),   
\end{equation} 
where \textit{d} is the shortest distance of a logit to the decision boundary. Mean() and Var() represent the mean and variance over all data samples, respectively. The purpose of normalizing with the variance is to compensate for the simple scaling of the logits. We invented this new metric (\textit{d${}_{s}$}) because it is impossible to calculate the Mahalanobis distance (\textit{d${}_{M}$}) as defined in \eqref{GrindEQ__2_} for a real-world dataset, in which the covariance matrices (\textbf{\textit{S}}) are different for each class.

We compare the \textit{d${}_{s}$} measured in our UniBERT with the one in BERT in Table 5. UniBERT's logits are much further away from the decision boundary compared to BERT's (i.e., Mean(\textit{d}) is 72.4 vs. 6.1); this increase is not the result of a simple scaling of the logits because the variance stays roughly the same (i.e., Var($d$) is 4.4 vs. 1.2). \textit{d${}_{s}$} succinctly summarizes the normalized distance to the decision boundary. We observe that \textit{d${}_{s}$} is tripled in UniBERT (16.4) than in BERT (5.0), which confirms our hypothesis that the multi-margin loss encourages distinct neural representations far away from the decision boundaries. Another way to interpret this data is that UniBERT increases the inter-class distance between the neural distributions for different classes while deterring the intra-class spread within the same class. As a consequence, with a larger buffer to absorb input perturbations, our UniBERT show superior post-attack accuracies as shown previously.

\begin{table}[!t]
\renewcommand{\arraystretch}{1.3}
\caption{\titlecap{Average distance to the decision boundary for the neural representations}}
\label{table_5}
\centering
\begin{tabularx}{\columnwidth}{p{0.4in}p{0.7in}*{4}{p{0.4in}}} \hline 
\textbf{Model} & \textbf{Loss} & \textbf{Unitary} & \textbf{Mean(\textit{d})} & \textbf{Var(\textit{d})} & \textbf{\textit{d${}_{s}$}} \\ \hline 
BERT & Cross-entropy & No & 6.1 & 1.2 & 5.0 \\\hline 
UniBERT & Multi-margin & Yes & 72.4 & 4.4 & 16.4 \\ \hline 

\end{tabularx}
\raggedright
\begin{footnotesize}
\newline
With the yelp dataset, we record the logits for 950 correctly classified samples and measure their shortest distances to the decision boundary.  For all 950 samples, Mean(\textit{d}) is the average distance, Var(\textit{d}) is the variance of \textit{d}, and \textit{d${}_{s}$} is the normalized distance (i.e., Mean(\textit{d})/ Var(\textit{d})). The UniBERT widens the distance between the neural representations of different classes while keeping the intra-class spread small, encouraging distinct representations for each class.
\end{footnotesize}
\end{table}

\subsection{Propagation of Perturbation}

\noindent With Theorem 1, we aim to use unitarity to regulate the magnitude of the perturbation. Nevertheless, as discussed in Sect. 4.2 previously, a completely unitary neural network is impossible because non-square weights and nonlinearities are needed for creating the model complexity required to solve complex problems. In our last experiment as follows, we demonstrate that, even with imperfect unitarity, our UniBERT can still stabilize the perturbations, preventing them from being scaled arbitrarily by the synaptic weights. 

To quantify the effect of a perturbation, we use the \textit{cosine similarity} to measure the alignment of two vectors representing the neural activations, which is defined as:
\begin{equation} \label{GrindEQ__9_} 
cosine\ similarity\triangleq \frac{\overline{a}\cdot\overline{b}}{{\left\|\overline{a}\right\|}_2{\left\|\overline{b}\right\|}_2}\boldsymbol{=}{\mathrm{cos} (\theta) \ },  
\end{equation} 
where $\overline{a}\ $and $\overline{b}$ are two vectors in the same vector space, $\cdot$ is the vector dot product, and \textit{$\theta$} is the angle between $\overline{a}\ $and $\overline{b}$. Cosine similarity ranges from -1 to 1, inclusive. When the two vectors overlap completely, the cosine similarity is the highest, signaling that the two vectors are identical. If they point in opposite directions, the cosine similarity is -1. In the context of NLP, researchers often the cosine similarity between the two corresponding sentence embeddings to measure their differences in semantics. 

Likewise, we use cosine similarity to quantify how much the activations deviate from their original values when the network is under attack. If the network is less susceptible to the input perturbations injected by the attacker, the attacked activations will be similar to their original, yielding a high cosine similarity. To visualize the neural network's susceptibility to injected perturbation, we measure the cosine similarity between the perturbed activations and the original activations across the neural net. Fig. 3 shows the results. In particular, we randomly select 1000 sentences from the ag\_news dataset, produce perturbed sentences using the Textfooler attack recipe, and record the activations after each of the 12 unit blocks in UniBERT (see Fig. 2 for the network architecture). We perform the same steps for BERT and compare the two results.

\begin{figure}[!t]
\centering
\includegraphics[width=3.5in]{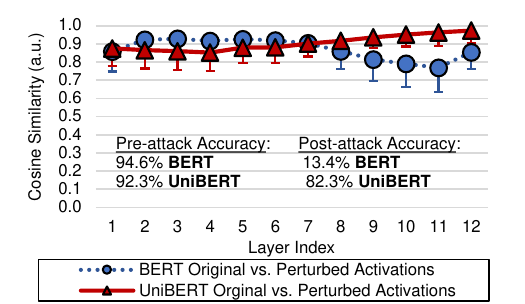}
\caption{Cosine similarity between the original and the perturbed activations for different neural networks. Adversaries attack the neural networks by perturbing the input sentence. We measure the cosine similarity between the activations of the original sentence and the activations of the perturbed sentence in UniBERT at the output of each attention layer (i.e., ${\overline{x}}_1\dots {\overline{x}}_{12}$ in Fig. 2) indexed from 1 to 12 across the neural network (dotted line). The same is done for BERT (dashed line). We perform statistical analysis on 1000 randomly chosen sentences from the ag\_news dataset and perturb them with the Textfooler attack recipe; the error bars denote standard deviation. As defined in \eqref{GrindEQ__9_}, cosine similarity ($\mathrm{\in }$$\mathrm{\mathbb{R}}$$\mathrm{\cap}$[-1,1]) is a distance measure for quantifying how well two vectors align with each other in a vector space with one indicating that the two vectors are identical. For UniBERT, the representations become more alike in deeper layers; on the contrary, the similarity fluctuates in BERT. A higher cosine similarity score means that the network is capable of restraining perturbations closer to the original neural representation; consequently, higher post-attack accuracies (see the text added in the figure).}
\label{fig_3}
\end{figure}

As shown by the dotted line in Fig. 3, BERT's cosine similarity fluctuates wildly across the different layers of the networks. In particular, it dips to the lowest value at layer 11 with a mean of 0.77 and a standard deviation of 0.14. The activations at layer 11 deviate from their original values significantly when BERT is under attack, and the cosine similarity quantifies this deviation. In short, when BERT is under attack, a significant portion of the sentences have activations that are a great deal different from their original values. On the other hand, our UniBERT fixes this problem by improving the similarity between the original and attacked activations (solid line, Fig. 3). UniBERT's cosine similarity is not only higher in the later layers but also varies less than BERT's as shown by the shorter error bars in the figure. At layer 11, UniBERT's cosine similarity is 0.96 with a standard deviation of 0.07, showing a significant reduction in variation compared to BERT's (i.e., 0.77$\mathrm{\pm}$0.14).

This result is interesting because only 62.3\% of the weights in UniBERT are unitary and there are many nonlinearities in the network (see Sect. 4.1). We argue that the success of UniBERT in stabilizing the perturbation comes from the unitarity added in the attention layers, which are repeated 12 times in the network. The attention mechanism is achieved primarily with the Key, Query, Value, and Dense modules in Fig. 2\textbf{;} the weights in these modules are all unitary in UniBERT as shown in Table 1. Although the decision processes in a neural net are generally non-interpretable, with the data in Fig. 3, we conclude that the novel unitary attention mechanism in our UniBERT encourages the network to make similar decisions in each layer even under adversarial attacks as shown with the rising cosine similarity of the activations shown in the figure for UniBERT (Fig. 3, solid line). 

It is difficult to exactly predict the post-attack accuracies from the cosine similarity between activations, but the two quantities are positively correlated. A higher cosine similarity means that the activations under attack are similar to the original ones; therefore, the network is more likely to achieve the pre-attack accuracies as if there is no attack because the logits are less likely to cross the decision boundary. For reference, we insert the pre-attack and post-attack accuracies in Fig. 3. UniBERT has higher cosine similarity compared to BERT in layers that are immediately preceding the final classifier (i.e., layers 8-12 in Fig. 3). The resilience to perturbation is especially important later in the network where the final decisions are made by demarcating the logits with decision boundaries. In summary, UniBERT delivers a superior post-attack accuracy (82.3\%) compared to BERT's (13.4\%) by keeping its activations closer to their original values.

\section{Conclusion}

\noindent Here we present an enhanced neural architecture named UniBERT for robust natural language processing. Our UniBERT defends against adversarial attacks, in which the adversaries attempt to slightly modify the input sentences to disrupt the prediction results in deep neural nets. It is a variant of BERT with two major improvements: First, we replace the cross-entropy loss with the multi-margin loss during finetuning to encourage distinct neural representations. Second, we force unitary constraints on the attention layer to restrain the perturbations injected during an attack. With the multi-margin loss and the unitary constraints, UniBERT is shown to be an effective defense technique against both typographical and synonym-based adversarial attacks, boosting the post-attack classification accuracy by more than 5.3\% compared to the state-of-the-art defense mechanisms across all datasets and attack recipes (Sect. 6.2). 

In addition, we study the contributions from multi-margin loss and unitarity in UniBERT individually, discovering that unitarity has to be used together with the multi-margin loss to be effective (Sect. 6.3) because it is necessary to first create distinct neural representations with the multi-margin loss (Sect. 6.4) to observe the effect of unitarity in post-attack accuracies. Unitary weights stabilize perturbations by keeping the activations close to their original values across the network (Sect. 6.5). Orthogonal to the state-of-the-art defense methods (e.g., adversarial training or regularization), our UniBERT is straightforward to implement and works well for a wide range of applications including categorization, natural language inferencing, and sentiment analysis. Our contributions empower deep learning practitioners to build computationally efficient and robust NLP pipelines for critical applications.

\ifCLASSOPTIONcompsoc
  \section*{Acknowledgments}
\else
  \section*{Acknowledgment}
\fi

The authors wish to thank Professor Robert N. Schwartz for his valuable discussions. This work was supported in part by the National Defense Science and Engineering Graduate Fellowship and is based on the first author's doctoral dissertation [26]. 

\ifCLASSOPTIONcaptionsoff
  \newpage
\fi


\nocite{*}
%




\end{document}


\Resetlcwords
\Addlcwords{for a is but and with of in as the etc on to if by}
\setcounter{figure}{3}
\setcounter{table}{5}

%
\appendices
\section{Training Details}
\noindent We document our training procedure for the proposed UniBERT. Training BERT models requires two phases: pre-training and finetuning. Pre-training teaches our UniBERT the basic mechanics of the language while finetuning provides on-the-job training for performing a specific task. Each phase requires a slightly different set of hyperparameters because of the differences in the task objectives (masked language modeling vs. classification) and the training datasets.

\begin{enumerate}
\item  \textbf{Pre-training:} We use a fixed language mask with a masking probability of 0.15 and pretrain the UniBERT model from scratch for five epochs with the Adam algorithm on a linear schedule, in which the learning rate will start with 0 at the beginning, linearly rise to 0.0001 (at step 7000), then linearly decay to zero (at the end of the training). Adam is a training algorithm with adaptive moment estimation that alleviates the problem of local minimum by adjusting the momentum of the weight updates. The weight decay reduces the magnitude of the synaptic weights to penalize against extravagant model complexity for better generalization performance. The batch size is 16; the sequence length is 512. These sizes are limited by the memory size of our graphics card where we run our simulation. Overall, the pretraining procedure takes 5 epochs to converge or equivalently 700,000 steps given our batch size and dataset size. At each backpropagation step, we update all synaptic weights with Adam and then convert selected ones to the closest unitary matrices using the QR factorization technique explained in Sect. 4.3. 

\item  \textbf{Finetuning:} We finetune the pretrained models on the three classification datasets: ag\_news, snli, and yelp. Similarly, we use QR factorization (Sect. 4.3) to ensure unitarity on selected weights. Unlike pretraining, the multi-margin loss is used instead of the cross-entropy loss. We set the margin parameter (\textit{$\varepsilon$}) to 100 for balancing the pre-attack and post-attack prediction accuracies. See Appendix B for details on selecting the best value for the margin parameter (\textit{$\varepsilon$}). We conduct four independent finetuning runs for scientific rigor; each has a new random initialization on the classifier layer. Finetuning takes five epochs with a learning rate of 0.00005. The batch size is 160 with a sequence length of 128; these settings are limited by the memory of the graphics card used in our simulation.
\end{enumerate}
\noindent The training procedures for the other baseline and defense models are left unchanged as described in their original publication. They use neither the unitary weights nor the multi-margin loss during any portion of the training process. We summarize the hyperparameters used for UniBERT training in Table 6 below. 

\begin{table}[!t]
\renewcommand{\arraystretch}{1.3}
\caption{\titlecap{Hyperparameters for the UniBERT training}}
\label{table_6}
\centering
\begin{tabularx}{\columnwidth}{XXXX} \hline 
\textbf{Parameter} & \textbf{Pretraining}  & \textbf{Finetuning} \\ \hline 
Model Type & Masked-Language & Classification \\ 
Loss Function & Cross-entropy & Multi-margin \\ 
Margin Parameter & N/A & 100 \\ 
Masking Ratio & 0.15 & N/A \\ 
Sequence Length & 512 & 128 \\ 
Unitary Constraint & Yes & Yes \\ \hline 
Dataset & bookcorpus & ag\_news, snli, yelp \\ 
Batch Size  & 16 & 128 \\ 
Epoch & 5 & 5 \\ \hline 
Optimizer  & Adam & Adam \\ 
Scheduler & Linear  & Linear \\
Learning rate  & 0.0001 & 0.00005 \\
Beta 1  & 0.9 & 0.9 \\ 
Beta 2 & 0.999 & 0.999 \\ 
Warmup Steps & 7000 & 500 \\ 
Weight Decay & 0.01 & 0.01 \\ \hline 
\end{tabularx}
\raggedright
\begin{footnotesize}
\newline
We first pretrain UniBERT and then finetune it with different datasets and loss functions shown in the table. These hyperparameters are the best-known methods reported in the literature except for the loss, margin, and unitary constraint, which are unique to UniBERT. In detail, the differences to train our UniBERT compared to a regular BERT are: First, UniBERT enforces unitary constraints on the selected layers listed in Table 1 for both pretraining and finetuning. Second, it uses the multi-margin loss with the margin parameter ($\varepsilon$) set to 100 for finetuning. The cross-entropy loss is used for pretraining. The learning rate starts with zero, linearly ramps up to the target learning rate at the prescribed warmup steps, and linearly ramps down to zeros. The sequence length reports the maximum number of words for each input. N/A means not applicable.
\end{footnotesize}
\end{table}

\section{Margin Parameter Selection}
\noindent There is a new hyperparameter in UniBERT: the margin parameter (\textit{$\varepsilon$}) for the multi-margin loss in (1). We find the best margin parameter, \textit{$\varepsilon$}, experimentally by sweeping it over five decades between 0.01 and 1000 to simultaneously maximize both the pre-attack and post-attack prediction accuracies of the neural net. We evaluate the tradeoff between pre-attack and post-attack accuracies for the yelp sentiment analysis in Fig. 4 below, in which our UniBERT is trained to classify the user text into positive or negative reviews. As the margin increases, the post-attack accuracy quickly improves (as shown by the dashed trendline in the figure) while the pre-attack accuracy reduced slightly with \textit{$\varepsilon$} (as illustrated by the dotted trendline). The best post-attack accuracy happens when \textit{$\varepsilon$} is 100; therefore, we use this setting for the rest of this paper. Depending on the user's tolerance for pre-attack accuracy drop in their applications, they can select an appropriate margin setting accordingly.

\begin{figure}[!t]
\centering
\includegraphics[width=3.5in]{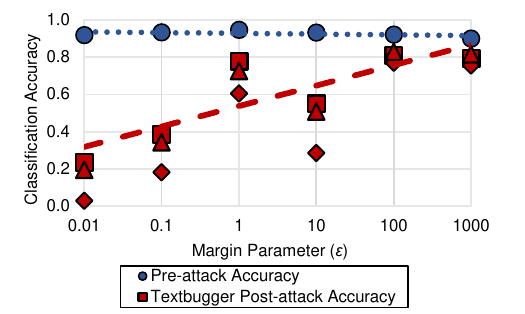}
\caption{Classification accuracy vs. margin parameter for the multi-margin loss in UniBERT. To find the most optimal setting for the margin parameter (\textit{$\varepsilon$}), we use the yelp sentiment analysis dataset to evaluate the pre-attack accuracy of UniBERT in making a correct classification, contrasting it with the post-attack accuracy under various adversarial attacks (namely, Textfooler, Textbugger, and PWWS). As shown in the figure, a higher $\varepsilon$ reduces the pre-attack accuracy (dotted line) while increasing the post-attack accuracy (dashed line) for UniBERT. From this data, we select \textit{$\varepsilon$} = 100 for the rest of the experiments in this treatise because, with this setting, the UniBERT delivers a large improvement in the post-attack accuracy with a negligible decline in the pre-attack accuracy.}
\label{fig_4}
\end{figure}

\section{Dataset Characteristics}
\noindent We highlight the key features of the dataset statistics in Table 7 below, including the number of classes for the human-generated labels, size of the training data set (training set), size of the testing dataset (test set), and average sample length. and all datasets used here follow a uniform distribution in the labels. Sentences are forced to be all lowercase if the dataset is cased, and all datasets are in English. The datasets are accessible for free from the Hugging Face repository [25].

\begin{table}[!t]
\renewcommand{\arraystretch}{1.3}
\caption{Key Features of the Dataset Statistics for the NLP Datasets}
\label{table_7}
\centering
\begin{tabularx}{\columnwidth}{XXXXX} \hline 
\textbf{Dataset} & \textbf{Number of Classes} & \textbf{Training Sample Size} & \textbf{Testing Sample Size}  & \textbf{Average Length\newline (Words)} \\ \hline 
bookcorpus & N/A & 70,000,000 & 4,000,000 & 13 \\ \hline 
ag\_news & 4 & 120,000 & 7,600 & 39 \\ \hline 
snli & 3 & 550,000 & 10,000 & 20 \\ \hline 
yelp & 2 & 560,000 & 38,000 & 136 \\ \hline 
\end{tabularx}
\raggedright
\begin{footnotesize}
\newline
bookcorpus is an unsupervised dataset used to train BERT and UniBERT as a mask language model during pretraining. ag\_news is a news classification dataset with four classes (world news, business news, science \& technology, and sports). snli is a natural language inferencing dataset that asks the model to predict the relationships between two sentences into three categories: entailment, contradiction, or neutral. yelp is a binary sentiment analysis dataset that categorizes user-generated paragraphs into positive reviews or negative reviews.  For bookcorpus, each sentence counts as a sample while the other three datasets can have multiple sentences per sample, and the average length column reports the average number of words in a sample. N/A means not applicable.
\end{footnotesize}
\end{table}